\documentclass[sigconf]{acmart}
\usepackage{dblfloatfix}
\usepackage{times}
\usepackage{soul}
\usepackage{url}
\usepackage[utf8]{inputenc}
\usepackage{graphicx}
\usepackage{amsmath}
\usepackage{amsthm}
\usepackage{booktabs}

\usepackage{algorithm}

\usepackage[switch]{lineno}
\usepackage{tabularx}
\usepackage{array}
\usepackage{multirow}

\usepackage{subcaption}
\usepackage{algpseudocode}
\usepackage{makecell}
\usepackage{url}
\usepackage{todonotes}
\usepackage{comment}

\makeatletter
\def\@copyrightpermission{}
\makeatother
\setcopyright{none}
\makeatletter
\def\@mkbibcitation{} 
\makeatother

\setcopyright{none} 
\acmDOI{} 
\acmISBN{} 
\acmBooktitle{} %

\AtBeginDocument{%
  }

\acmConference[Pubished in ICAIF Workshops '24]{5th ACM International Conference on AI in Finance}{Nov 14--16,
  2024}{Brooklyn, NY}

\begin{document}

\title[Evaluating Financial Sentiment Analysis with Annotators’ Instruction Assisted Prompting]{Evaluating Financial Sentiment Analysis with Annotators’ Instruction Assisted Prompting: Enhancing Contextual Interpretation and Stock Prediction Accuracy}

\author{A M Muntasir Rahman}
\email{ar238@njit.edu}
\affiliation{%
  \institution{Department of Computer Science\\New Jersey Institute of Technology}
  \city{Newark}
  \state{New Jersey}
  \country{USA}
}

\author{Ajim Uddin}
\email{ajim.uddin@njit.edu}
\affiliation{%
  \institution{Martin Tuchman School of Management\\New Jersey Institute of Technology}
  \city{Newark}
  \state{New Jersey}
  \country{USA}
}

\author{Guiling ``Grace” Wang}
\email{gwang@njit.edu}
\affiliation{%
  \institution{Department of Computer Science\\New Jersey Institute of Technology}
  \city{Newark}
  \state{New Jersey}
  \country{USA}
}









\begin{abstract}
  Financial sentiment analysis (FSA) presents unique challenges to Large Language Models (LLMs) that surpass those in typical sentiment analysis due to the nuanced language used in financial contexts. The prowess of these models is often undermined by the inherent subjectivity of sentiment classifications in existing benchmark datasets like FiQA and Financial Phrasebank. These datasets typically feature undefined sentiment classes that reflect the highly individualized perspectives of annotators, leading to significant variability in annotations. This variability results in an unfair expectation for LLMs during benchmarking, where they are tasked to conjecture the subjective viewpoints of human annotators without sufficient context. In this paper, we introduce the Annotators’ Instruction Assisted Prompt (AIAP), a novel evaluation prompt designed to redefine the task definition of FSA for LLMs. By integrating detailed task instructions originally intended for human annotators into the LLMs' prompt framework, AIAP aims to standardize the understanding of sentiment across both human and machine interpretations, providing a fair and context-rich foundation for sentiment analysis. We utilize a new dataset (WSBS) derived from the WallStreetBets subreddit to demonstrate how AIAP significantly enhances LLM performance by aligning machine operations with the refined task definitions. Experimental results demonstrate that AIAP enhances LLM performance significantly, with improvements up to 9.08\%. This context-aware approach not only yields incremental gains in model performance but also introduces an innovative sentiment-indexing method utilizing model confidence scores. This method enhances stock price prediction models and extracts more value from the financial sentiment analysis, underscoring the significance of WallStreetBets as a critical source of financial text. Our dual-pronged research offers insights into both improving FSA through better evaluation methods and applying these advancements in practical security analysis. WSBS Dataset available at \textbf{\url{https://github.com/moonscape95/WSBS}}

\end{abstract}


\keywords{Financial Sentiment Analysis, Large Language Model, Prompt Design, WallStreetBets}


\maketitle

\section{Introduction}

The domain of financial sentiment analysis (FSA) poses considerable challenges, specifically when employing Large Language Models (LLMs) due to the specialized linguistic nuances prevalent in financial texts. These challenges are compounded by the subjectivity inherent in sentiment classifications found in existing benchmark datasets such as Financial Phrasebank \cite{malo2014good}, FiQA-SA \cite{FIQA}, and the Twitter Financial News Sentiment Dataset \cite{twitterfns}. These datasets often suffer from a lack of standardization in sentiment categories, reflecting the diverse interpretations of individual annotators, which introduces substantial variability in annotations. This variability sets unrealistic benchmarks for LLMs, as they are expected to infer the subjective judgments of human annotators without adequate context.

In typical sentiment analysis tasks, models are often prompted with oversimplified queries like “What is the sentiment of this sentence?”, which do not provide enough guidance on the nuanced differences between sentiment categories. As a result, LLMs are forced to infer a generalized understanding of positive, negative, and neutral sentiments within the complex financial domain, adding a significant layer of complexity to their task.

To mitigate the unfair evaluation of LLMs on benchmarking datasets that hold subjectively labeled samples, this paper introduces the Annotators’ Instruction Assisted Prompt (AIAP), a novel evaluation prompt that incorporates detailed task instructions originally intended for human annotators into the LLMs' prompt structure. The AIAP is designed to standardize the understanding of sentiment across both human and machine interpretations, providing a consistent and context-rich basis for sentiment analysis. By clearly defining what each sentiment class means and ensuring that both annotators and models adhere to these definitions, AIAP aims to lessen the ambiguity that typically disrupts the fair performance-evaluation of LLMs in FSA tasks.

Employing a novel dataset derived from the WallStreetBets subreddit, we demonstrate that when equipped with the same task definitions as human annotators, LLMs such as GPT-4 \cite{openai2024gpt4technicalreport}, Llama-3 \cite{dubey2024llama3herdmodels}, FinGPT \cite{yang2023fingpt} gains accuracy boosts of 5.90\% on average, and 9.08\% on the best case scenario. This novel dataset and the accuracy improvement gives us two key insights. First, annotators' subjectivity on financial sentiment classes can converge with a simple set of instructions. Second, LLMs can align their inference with human annotators using the same set of instructions by taking advantage of their in-context learning capacity.

\begin{figure}[htbp]
\centering
\includegraphics[width=1.0\columnwidth]{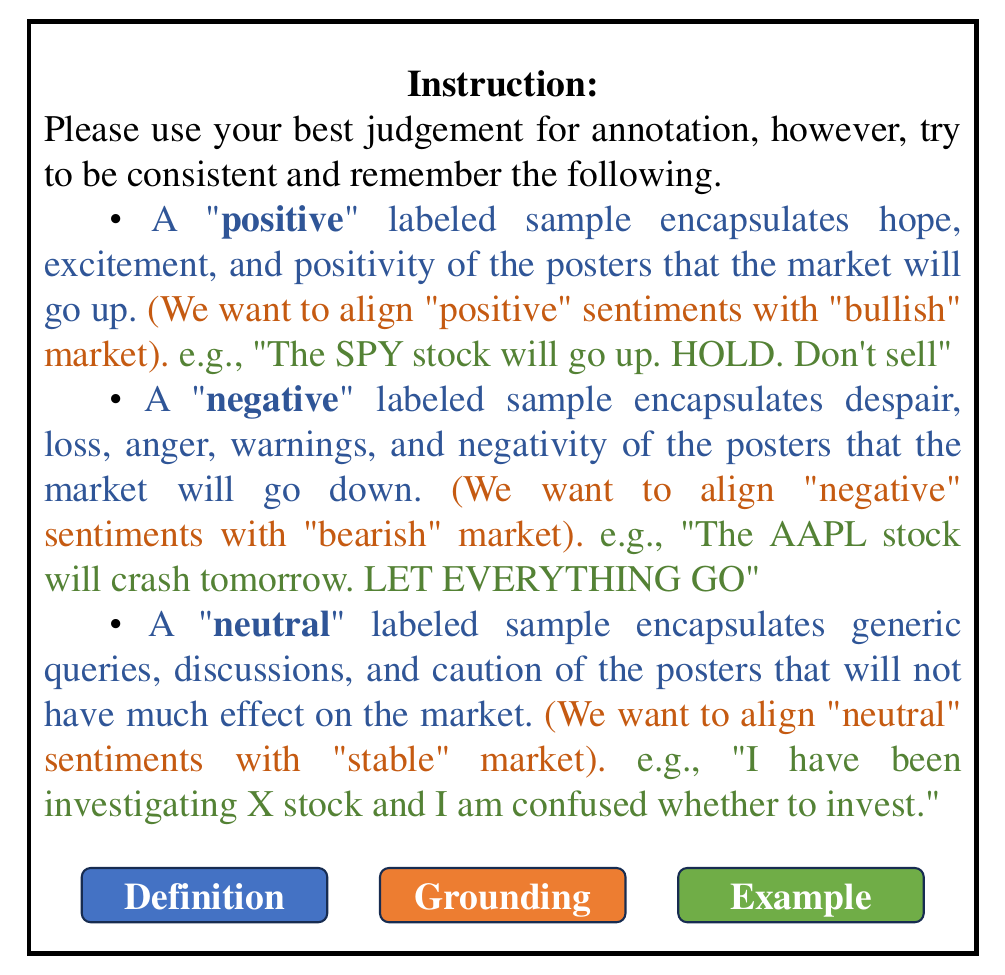} 
\caption{Instruction for the annotators of the WallStreetBets Sentiment Analysis Dataset.}.
\label{fig:annotators_instruction}
\end{figure}

The selection of WSB (r/WallStreetBets \footnote{\url{https://www.reddit.com/r/wallstreetbets/}}) as a source for financial text data in our dataset is motivated by several factors. First, its significant popularity following the 2021 GameStop (GME) short squeeze event. Second, the relevance of financial sentiment analysis is heightened when examining its impact on the market. The influence of social media sentiment, especially of a platform like WSB, is of particular interest at the time being. To assess the impact of WSB's sentiment on the stock market, we conduct some stock prediction analysis as well. Aside from FinGPT, we also fine-tune a transformer-encoder based classifier on our annotated dataset and analyze daily posts/comments to derive sentiment indices for specific stocks. These indices are then incorporated as additional features into stock price prediction models. We introduce a sentiment-score generation method based on the confidence score generated by the encoder-based models, CSBS, during sentiment classification. Our experimental results indicate that CSBS can extract more value from the encoder models during stock prediction using our annotated dataset.

Overall, in this study we try to first improve financial sentiment analysis by in-context learning and then focus on applying financial sentiment analysis using quantitative and qualitative approaches. The following key points can summarize our contributions.
\begin{itemize}
    \item We introduce a novel expert-annotated dataset for financial sentiment analysis with well defined annotator instructions for fair benchmarking.
    \item We examine the importance of aligning human annotators' and LLMs' understanding of the task of financial sentiment analysis by introducing a better-structured prompt.
    \item We empirically demonstrate that the more context-aware the sentiment analysis is, the better the LLMs perform.
    \item We propose using sentiment classifier confidence scores to enhance stock price prediction.
\end{itemize}
 
The paper is organized as follows: section \ref{section: Dataset} discusses the dataset and explains AIAP, section \ref{experimental_setup} details experimental setup, while section \ref{experimental_results} explains experimental results, section \ref{section: stock prediction} talks about stock prediction applications, section \ref{section: relevant studies} organizes the relevant works, and section \ref{section: conclusion} concludes the paper. We will release our dataset and code to the public upon acceptance.
\section{Annotating WallStreetBets' Sentiment}
\label{section: Dataset}

Humans have preconceived notions about the basic financial sentiment classes (positive/negative/neutral). As a result, researchers observe disagreements among the annotators of financial sentiment analysis datasets (e.g., in Financial Phrase Bank, the annotators had 100\% agreement over less than 50\% of the entire dataset \cite{malo2014good}). Even with precise definitions and examples, it can be challenging to expect two different human beings to have identical thought processes. However, some guidelines can definitely be provided to the annotators to nudge them towards a common definition and idea of what constitutes the basic sentiment classes for the financial industry. With this objective, we provide a brief instruction with three key components to our annotators as a compass so that their notion of sentiment does not diverge too much.
\begin{itemize}
    \item Definition: We provide a core definition of each label in the financial context using synonymous terms that help describe the intangible concept/feeling.
    \item Grounding: The basic sentiment classes (positive, negative, neutral) are equivalently used for a different set of labels grounded in financial text (bullish/bearish/stable) frequently (e.g., Twitter Finance Sentiment Dataset). We add a grounding component to help align these two sets of labels.
    \item Example: We provide one unambiguous example for each class alongside the definitions.
\end{itemize}
Figure \ref{fig:annotators_instruction} color-codes these components and presents the entire instruction.

\subsection{Annotation Process \& Annotators' Agreement}
To ensure a balanced distribution of samples with varying sentiment tones, a collection of 3,000 samples (posts \& comments) were
carefully handpicked from WallStreetBets. We chose five graduate-level researchers, proficient in financial terminology and acquainted with the informal tone and colloquial language used in WSB, to serve as annotators. The annotation was completed in two stages. In the initial stage, every sample received two different annotations from different annotators. In the final stage, two annotators re-examined each sample, and conflicts were resolved by majority voting. The annotators were each given 500-1000 samples to annotate within three weeks in the first round.

This annotation process produced 2,920 sentimentally relevant samples; 80 were marked as irrelevant and discarded. Out of these 2920 samples, 1509 had unanimous annotations in the first stage of annotation and the rest of the samples were labeled according to the conflict resolution stage. This essentially produced two sets of our WallStreetBets-Sentiment Analysis (WSBS) dataset: Full Dataset, All-Agree Dataset. The class distribution of these sets are presented in table \ref{table:dataset_stats}.

\begin{table}[!htbp]
\centering
\caption{WallStreetBets Sentiment (WSBS) dataset statistics}
\label{table:dataset_stats}
\begin{tabular}{ccccc}

\toprule
Dataset & Positive & Negative & Neutral & Total  \\ \midrule
All-Agree & 43.2\% & 33\% & 23.8\% & 1509\\ 
Full  & 38.8\% & 29.3\% & 31.8\% & 2920 \\ 
                        \bottomrule
\end{tabular}

\end{table}

\subsection{Annotators' Instruction Assisted Prompt} 

\label{section: AIAP}
We want to test if a frozen LLM can merge its sentiment prediction with the annotators if provided with the same guidelines. This approach helps us understand the complexity of financial sentiment analysis and improve model prompts during evaluation from simple tertiary classification to a context-aware sentiment classification. For the AIAP, we simply insert the instruction provided to the annotators shown in figure \ref{fig:annotators_instruction} into the base prompt in the following manner.


\begin{itemize}
\renewcommand\labelitemi{}
    \item \textbf{Instruction: }What is the sentiment of this 
\textbf{input}? Please choose an answer from {negative/neutral/positive}
    \item $\textless$ Insert Annotators' Instruction Here $\textgreater$
    \item \textbf{Input: }Financial terms were not disclosed.
    \item \textbf{Answer: }neutral
\end{itemize}

Here, the bold faced token ``\textbf{input}" is the identifier term as presented later in table \ref{table:AIAP_performance}. We used three different identifier terms: ``input", ``news", and ``tweet". We observed varying performances by changing just this single term in the prompt, hence, we performed all our experiments using these different ``identifier" terms.




\section{Experimental Setup}
\label{experimental_setup}
The key experiments for this work is relied upon two major components: Models and Prompts. The model selection here is crucial as we want to observe in-context learning ability of LLMs. On the other hand, prompt designing is one of the most significant contributions of the paper.
\subsection{Models}
As we are concerned with understanding instructions for language models, we look for instruction-tuned state-of-the-art LLMs. Instruction-tuning is the process of fine-tuning an LLM with instructions to complete different natural language processing tasks.The reasoning behind this is to distinguish the next-token-prediction pre-training objective from the realistic task objective so that models can perform a task without adapting to the task's structure and syntax. To ensure diversity, we selected one financial model, one open-source generic LLM, and one closed-source generic LLM. To this end, we utilized the following models for our experiments:
\begin{itemize}
    \item FinGPT: We selected FinGPT as one of our experimental model as it has instruction-tuned variants for sentiment analysis called FinGPT-SA \footnote{\url{https://huggingface.co/FinGPT/fingpt-sentiment_internlm-20b_lora}}. It is a 20 Billion parameter model based on the InternLM-20B model \cite{team2023internlm}
    \item Llama3: Llama3 \cite{dubey2024llama3herdmodels} is one of the largest open-source state-of-the-arts model available. We used the instruction tuned version of Llama3 with 70 billion parameters \footnote{\url{https://huggingface.co/meta-llama/Meta-Llama-3-70B-Instruct}}.
    \item GPT-4: GPT-4 \cite{openai2024gpt4technicalreport} is the pioneering and the most popular instruction-tuned model at present with 1.76 trillion parameters. We used the latest GPT-4 Omni \footnote{\url{https://openai.com/index/hello-gpt-4o/}} for our experiments. \end{itemize}

\subsection{Prompts}
We used FinGPT-SA, Llama3, and GPT-4 for our experiments on both All-Agree and Full sets of WSBS. The models are kept in inference mode and only the prompt is changed. The ``Base-Prompt" for all the models are represented in the following fashion:
\begin{itemize}
\renewcommand\labelitemi{}
    \item \textbf{Instruction: }What is the sentiment of this 
\textbf{input}? Please choose an answer from {negative/neutral/positive}
    \item \textbf{Input: } < Sample from WSBS to be classified into the sentiment classes >
    \item \textbf{Answer: } < Model generates response >
\end{itemize}
The ``AIAP" for all the models are represented in the following fashion:
\begin{itemize}
\renewcommand\labelitemi{}
    \item \textbf{Instruction: }What is the sentiment of this 
\textbf{input}? Please choose an answer from {negative/neutral/positive}
    \item $\textless$ Insert Annotators' Instruction from figure  \ref{fig:annotators_instruction} Here $\textgreater$
    \item \textbf{Input: } < Sample from WSBS to be classified into the sentiment classes >
    \item \textbf{Answer: } < Model generates response >
\end{itemize}

As there is no fine-tuning involved, the models only rely on the additional guidance laid out for them in the Annotators' instruction added in AIAP to garner more knowledge about the nature of the samples and response accordingly. For both Base-prompt and AIAP we run the experiments separately for the three different sample-identifier terms as explained in section \ref{section: AIAP}; this adds to the robustness of our experiments ensuring that the performance fluctuations are not some random signals.

\begin{table}[b]
\caption{Accuracy comparison between Base-prompt and AIAP on FinGPT-SA, Meta-Llama3-8B-Instruct, and GPT-4 in inference mode. All-Agree and Full are the two different sets of the WSBS dataset. ``Identifier" terms are explained in section \ref{section: AIAP}}
\centering
\begin{tabular}{cccccc}
\hline
\multirow{2}{*}{\rule{0pt}{2.5ex}Model} &\multirow{2}{*}{\rule{0pt}{2.5ex}Dataset} & \multirow{2}{*}{\rule{0pt}{2.5ex}Identifier} & \multicolumn{2}{c}{\rule{0pt}{2.5ex}Prompt} &\multirow{2}{*}{\rule{0pt}{2.5ex}Gain}\rule[-1.2ex]{0pt}{0pt} \\\cline{4-5}
                 &      &                        & \rule{0pt}{2.5ex}Base-Prompt& \rule{0pt}{2.5ex}AIAP & \\ \midrule
\multirow{6}{*}{FinGPT} & \multirow{3}{*}{All-Agree} & news & 64.74 & 72.37 & +7.63 \\ 
                       &                          & tweet & 56.53 & 62.23 & +5.70 \\  
                       &                          & input & 58.85 & 67.93 & +9.08 \\ \cline{2-6}
                           & \multirow{3}{*}{Full} & news & 57.19 & 61.30 & +4.11 \\ 
                       &                          & tweet & 51.44 & 57.91 & +6.47 \\  
                       &                          & input & 53.53 & 59.28 & +5.75 \\ \midrule
\multirow{6}{*}{Llama3} & \multirow{3}{*}{All-Agree} & news & 60.77 & 68.33 & +7.56 \\ 
                       &                          & tweet & 65.81 & 72.50 & +6.69 \\  
                       &                          & input & 63.22 & 67.46 & +4.24 \\ \cline{2-6}
                           & \multirow{3}{*}{Full} & news & 52.81 & 57.84 & +5.03 \\ 
                       &                          & tweet & 56.85 & 61.00 & +4.15 \\  
                       &                          & input & 53.87 & 56.34 & +2.47 \\ \hline
\multirow{6}{*}{GPT-4} & \multirow{3}{*}{All-Agree} & news & 70.25 & 79.06 & +8.81 \\ 
                       &                          & tweet & 73.76 & 78.76 & +5.00 \\  
                       &                          & input & 74.95 & 80.91 & +5.96 \\ \cline{2-6}
                           & \multirow{3}{*}{Full} & news & 58.39 & 65.38 & +6.99 \\ 
                       &                          & tweet & 60.75  & 63.80 & +3.05 \\  
                       &                          & input & 60.96 & 68.39 & +7.43 \\ \hline
\end{tabular}

\label{table:AIAP_performance}
\end{table}

\section{Experimental Results}
\label{experimental_results}
Table \ref{table:AIAP_performance} details the performance comparisons we set up in the earlier section. We can observe that on average the models exhibit 5.90\% accuracy improvement, and on the best case scenario a noteworthy 9.08\% accuracy improvement when equipped with the extra guidance, instead of simply asking to classify among positive/negative/neutral. Additionally, the best performance with the base-prompt is 74.95\% and the best performance with the AIAP is 80.91\%. It is evident that when prompted with the additional instruction, the LLMs perform consistently and significantly better compared to the base-prompt.
\begin{figure*}[htbp]
    \centering
    \begin{subfigure}[b]{0.45\textwidth}
        \includegraphics[width=\textwidth]{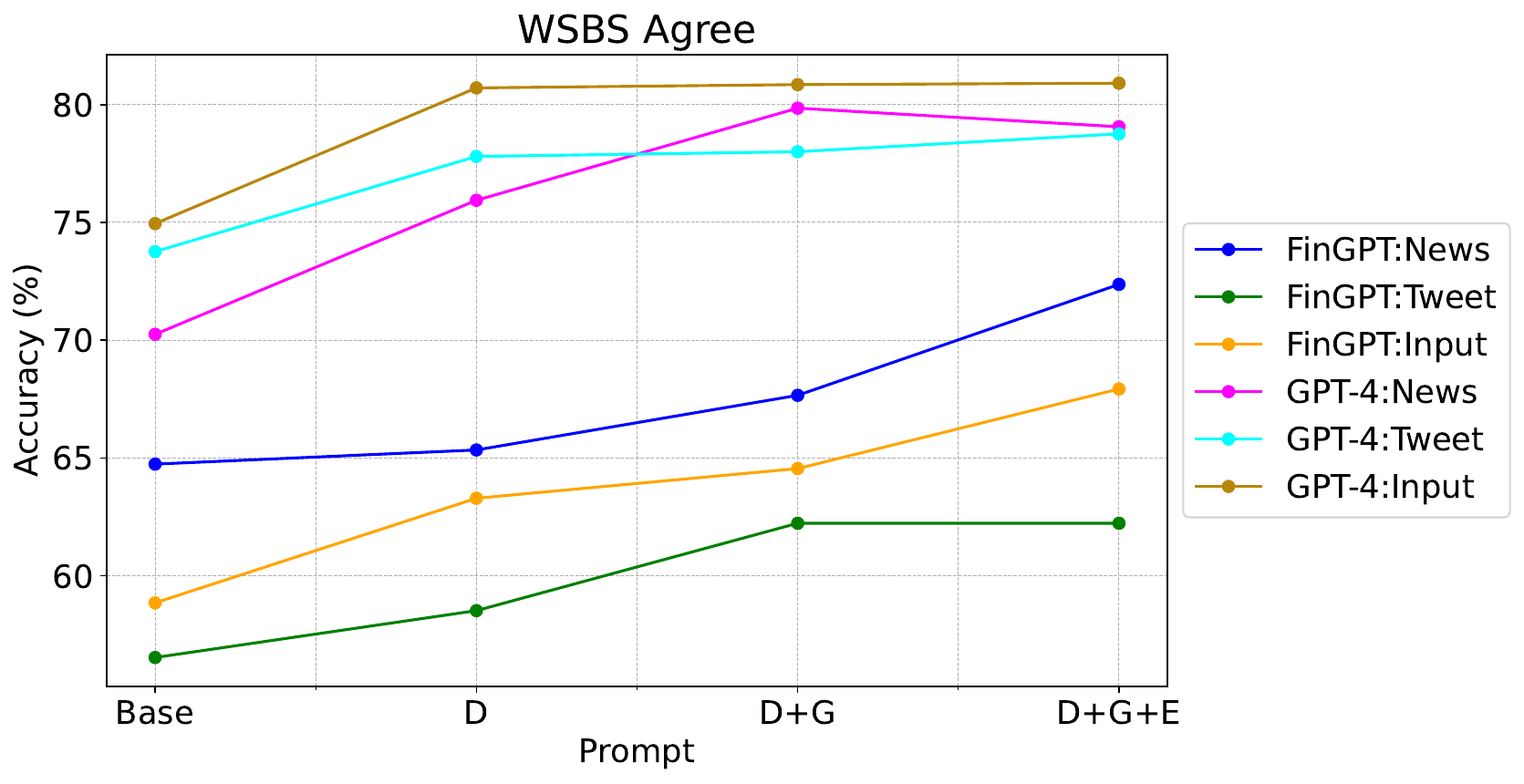}
        \label{subfig:1}
    \end{subfigure}
    \begin{subfigure}[b]{0.45\textwidth}
        \includegraphics[width=\textwidth]{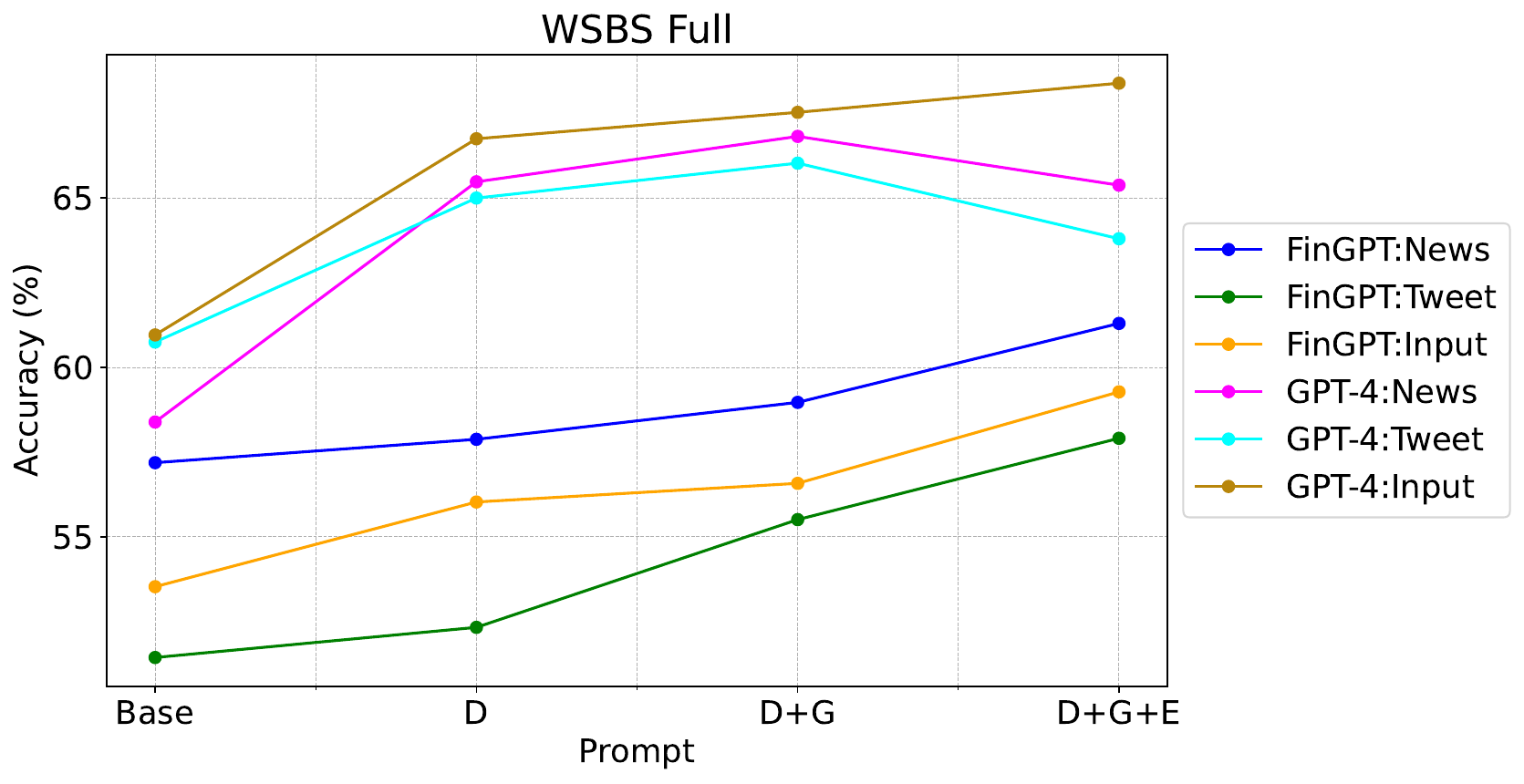}
        \label{subfig:3}
    \end{subfigure}
    \vspace{-0.2in}
    \caption{Better instruction-alignment or context-awareness ensures better outcome. Gradual performance increase as LLMs have access to more and more components of the instruction for the annotators. Y-axis represents accuracy and X-axis represents the prompts' content. D = Base Prompt + Definition added, D+G = D + Grounding added, D+G+E (Full AIAP) = D+G + Example Added.}
    \label{figure: better_instruction_ensure}
\end{figure*}

\subsection{Better Instructions Ensure Better Outcomes}

To test the robustness of the proposed prompting method, in other words, to validate that the performance gains are not some random noises and that there is some depth to the gain, we ask the following question.
\begin{itemize}
\renewcommand\labelitemi{}
    \item {\it Does the model's performance gradually increase as the instruction gets better and more aligned with the annotators?}
\end{itemize}
\textbf{Essentially, we hypothesize that the model's performance should gradually decrease if it does not have access to the entire instruction.} To test this, we resort to the three-component structure in our instruction (figure \ref{fig:annotators_instruction}). We check FinGPT-SA's accuracy on WSBS on three altered versions of the AIAP.

\begin{itemize}
    \item \textbf{Definition: } AIAP has just the core-definition (Blue fonts in figure \ref{fig:annotators_instruction})
    \item \textbf{Definition + Grounding: } AIAP has only the core definition and grounding direction (Blue and Orange fonts in figure \ref{fig:annotators_instruction}).
    \item \textbf{Definition + Grounding + Example: } AIAP has the whole instruction (All in figure \ref{fig:annotators_instruction})
\end{itemize}

In figure \ref{figure: better_instruction_ensure}, we showcase the performance improvements of FinGPT and GPT-4 on both sets of WSBS and the three different sample identifiers of the base prompt using the prompt variations of AIAP. In the x-axis, we first look at the base-prompt, and then we add just the ``definition" components of AIAP. As we move up the x-axis, we add the last two components, essentially providing more context to the LLMs. We observe that in majority of the cases, for each additional component added to the prompt, the performance gradually increases and does not fluctuate. This confirms our assumption that LLMs' performance indeed increases step by step as the instruction becomes more enhanced and aligned with the annotators'. However, the drops in 2 out of the 6 cases from ``D + G" to ``D + G + E" in the WSB-Full dataset could be attributed to either the ambiguity of the annotations in the samples of the full dataset, or the potential over-specific nature of the ``examples" provided in the prompt, as prompt-wise the examples are the only difference. Future study can focus on the nature of the ``examples" provided in the prompt.

\begin{table}[!htbp]
\caption{Accuracy comparison between base-prompt, few-shot prompts, and AIAP on FinGPT-SA}
\centering
\begin{tabular}{ccccccc}
\toprule
\multirow{2}{*}{\rule{0pt}{2.5ex}Dataset} & \multicolumn{5}{c}{\rule{0pt}{2.5ex}Prompt}\rule[-1.2ex]{0pt}{0pt} \\\cline{2-6}
                 & \rule{0pt}{2.5ex}BP& \rule{0pt}{2.5ex}1-S& \rule{0pt}{2.5ex}2-S& \rule{0pt}{2.5ex}3-S & \rule{0pt}{2.5ex}AIAP \\ \midrule
All-Agree& 64.74 & 65.94 & 58.58 & 62.82 & 72.37\\ 
Full  & 57.19 & 59.18 & 54.59 & 58.77 & 61.30  \\ 
                        \bottomrule
\end{tabular}

\label{table:few_shot_experiment}
\end{table}

\subsection{Comparison with Few-Shot Prompting}

Even though we use three examples in AIAP, it is not in the similar fashion as typical few-shot prompting. Few-shot prompting has been known to be successful in guiding LLMs toward a better understanding of the task \cite{brown2020language}. To examine whether FinGPT simply requires few-shot prompting to achieve significant improvement, we compare AIAP with one, two, and three-shot prompting (1-S, 2-S, 3-S respectively) using \textit{Instruction: Input: Answer: Input: Answer:} format and present in table \ref{table:few_shot_experiment}. We used the same three examples we used in AIAP and picked ``news" as the sample identifier as it performed the best on all the other experiments. It is quite evident from the table that few-shot prompting cannot improve performance (sometimes randomly degrades) as AIAP. This demonstrates that while examples in instructions are important, they serve better if structured well with the definition of the task (Figure \ref{figure: better_instruction_ensure} shows us the best gain when the examples were added to AIAP).

\subsection{Is this a Lottery Prompt?}
In prompt engineering for LLMs, researchers have noted the impact of Lottery/Magic/Secret prompts, the full extent of which remains to be comprehensively understood \cite{chen2023exploring}. For instance, in \cite{li2023starcoder} the authors present a prompt that seems only effective for one model but across multiple datasets. Inferring that certain prompts can elicit better model performance, regardless of the context. To infer that AIAP for WSBS is not a ``lottery prompt", it should exhibit two behaviors. First, it should work well on a different model with WSBS. Second, it should not work very well on a different dataset with FinGPT-SA, as AIAP in figure \ref{fig:annotators_instruction} is dataset-specific. We could confirm the former by testing on different models, observing on average 5.90\% improvement. The latter was confirmed by applying the same AIAP on Financial Phrase Bank (Test Set). We observed that it could only improve FinGPT's performance from 86.0\% to 86.4\% in the best case scenario. Considering the AIAP (particularly the examples) are not meant for FPB, the low and unstable performance is logical.

\section{Application in Stock Price Prediction}
\label{section: stock prediction}


Financial sentiment analysis can only take us so far; to fully utilize its potential, we must also analyze its application. As WSB is a platform heavily focused on the stock market and driving a herding behavior, we chose to analyze its sentimental impact on stock price prediction. We selected the top 5 most discussed stock tickers in WSB for our analysis. To be precise, we look at day-to-day posts/comments of the users of WSB, use AIAP on FinGPT-SA to predict the sentiment of these texts, generate a sentiment score for each day for each stock, and use this score as an added feature on regression models for stock price prediction to see if the sentiment feature can improve the models' prediction. We also wanted to take advantage of our WSBS dataset as a fine-tuning dataset for this experiment and analyze some small-scale transformer-based language models for sentiment prediction. We develop a novel confidence-score-based sentiment scoring method (CSBS) to extract more value out of these small-scaled classifiers fine-tuned on our annotated dataset for stock price prediction.

\subsection{Fine-tuning on WSBS}
We initially focused on finding the best-performing classifier on WSBS. To achieve this, we selected a few transformer-encoder-based \cite{araci2019finbert,devlin2018bert,distilbert} and transformer-decoder-based models \cite{OPT_paper,radford2019languageGPT}. Transformer-based models typically consist of two different building blocks: encoders and decoders. Classifier models are constructed by stacking the encoder modules of Transformers, while generative models are built using the decoder modules. Notably, generative models can be repurposed for classification tasks. The fine-tuning task was fairly straightforward. Each model had to classify the text into one of the three classes (Positive/Negative/Neutral). A 10-fold cross-validation was performed on the annotated dataset with a 90:10 train-to-test split. The ``bert-base-uncased" model was ultimately chosen for sentiment score generation as it performed the best on average among all the tested models. Details are presented in table \ref{tab:fine_tune_model_performance}. Although it is not an apple-to-apple comparison, it needs to be noted that without any fine-tuning, FinGPT-SA performed better with AIAP than fine-tuned classifiers, especially FinBERT.
\begin{table}
    \centering
    \caption{Fine-tuning performances with different models on WSBS. Numbers represent average over 10-fold cross-validation.}
    \label{tab:fine_tune_model_performance}
    \begin{tabular}{cc}
    \toprule
    \textbf{Model} & \textbf{Accuracy (\%)} \\
    \midrule
    Bert-base-cased & 68.20\\
 
    Bert-base-uncased & \textbf{70.30} \\

    Distilbert-base-cased & 69.10 \\

    Distilbert-base-uncased & 68.40 \\

    GPT-2 small & 66.90 \\
   
    GPT-2 medium & 68.20 \\

    OPT-125M & 59.50 \\

    FinBERT & 67.50 \\
    \bottomrule
    \end{tabular}
    
\end{table}

\textbf{FinBERT performs worse than BERT-base:} The experiments resulted in a keen observation that FinBERT, despite being a finance-specific language model failed to perform better than its base variant on the WSBS dataset. This observation can be attributed to several factors:

\begin{enumerate}
  \item \textbf{Domain Mismatch}
  \begin{itemize}
    \item \textbf{FinBERT's Training Domain:} FinBERT is pre-trained on financial texts such as SEC filings, earnings calls, and analyst reports. These texts are typically formal and structured, with financial jargon \cite{araci2019finbert}.
    \item \textbf{WallStreetBets Texts:} The WallStreetBets subreddit often features informal, slang-heavy language, memes, sarcasm, and insider jargon unique to that community. This style is vastly different from FinBERT's training domain, leading to a domain mismatch.
  \end{itemize}

  \item \textbf{Vocabulary and Tokenization}
  \begin{itemize}
    \item The tokenizer used for FinBERT is optimized for formal financial language and struggles with slang, abbreviations, and unique WallStreetBets terminology. In contrast, BERT-base, which is more general-purpose and pre-trained on a broader corpus including informal texts, handles these elements more effectively.
  \end{itemize}

  \item \textbf{Over-Specialization of FinBERT}
  \begin{itemize}
    \item FinBERT's pretraining makes it overly specialized for financial tasks requiring analysis of structured financial reports or traditional financial sentiment. This specialization may make it less adaptable to tasks involving non-standard language, sarcasm, and cultural nuances.
  \end{itemize}

  \item \textbf{BERT-Base's Generalization}
  \begin{itemize}
    \item BERT-base, being pre-trained on a large, diverse corpus (Wikipedia and BooksCorpus), has a more general understanding of language. It may perform better on datasets with informal, mixed-domain language like WallStreetBets posts.
  \end{itemize}
\end{enumerate}

\subsection{Sentiment Score Generation}
\label{section:sentiment_score_generation}

The task of generating a sentiment score for a specific timeframe can be described as follows: Collect relevant texts from the given timeframe, determine their sentiment predictions, and then compute a single sentiment score using a predefined function. It can be formalized as,
\[
\text{Sentiment Score} = f(\text{sent}_1, \text{sent}_2, \ldots, \text{sent}_n)
\]
Here, the function \( f \) represents the predefined function that aggregates individual sentiment predictions \( \text{sent}_1, \text{sent}_2, \ldots, \text{sent}_n \) from the collected texts into a single sentiment score. In a relevant study Hiew et al. proposed a quantitative sentiment index generation equation, and we used this equation as one of the sentiment score generation method for our experiments (referred to as QuantSS in this paper) \shortcite{QuantSS}. The score for stock $s$ on day $d$ can be calculated through:

\begin{equation}
\label{eq:baseline_sentiment_score}
    QuantitativeScore(s, d) = \frac{Pos_s^d - Neg_s^d}{Pos_s^d + Neu_s^d + Neg_s^d} 
\end{equation}

Where $Pos_s^d, Neu_s^d, Neg_s^d$ are the respective quantities of positive, neutral, and negative samples for stock $s$ on day $d$.

While QuantSS sentiment score generation method extracts a quantitative representation of each day's score, we devised a qualitative method to gather more insights from the posts and comments with the classifiers. Our approach uses the confidence scores (prediction class probability distribution) from the classifiers for the predicted class for each sample and decides the polarity of the neutral class based on the polarity of the other two classes. Note that this approach is confidence-score based, only applies to classification models, and cannot be used with generative models such as FinGPT. Algorithm \ref{algorithm} presents this scoring scheme with a pseudocode.
\begin{algorithm}[htbp]
\caption{Sentiment Analysis Scoring Algorithm}
\begin{algorithmic}[1]
\State \textbf{Input:} Daily comments and posts.
\State \textbf{Output:} Sentiment Score (CSBS) per day. 
\For{each day}
    \State $Score \gets 0$
    \For{each comment or post}
        \State $Logits \gets \text{getLogits}(\text{comment/post})$
        \State $P \gets \text{getSoftmaxProbability}(Logits)$
        \If{highest prob. is Positive}
            \State $Score \mathrel{{+}{=}} P[\text{pos}]$
        \ElsIf{highest prob. is Negative}
            \State $Score \mathrel{{-}{=}} P[\text{neg}]$
        \Else
            \State $Score \mathrel{{+}{=}} \text{sign}(P[\text{pos}] - P[\text{neg}]) \cdot P[\text{neu}]$
        \EndIf
    \EndFor
\EndFor
\end{algorithmic}
\label{algorithm}
\end{algorithm}



\subsection{Stock Prediction Experimental Setup}
For the stock prediction regression experiments, we use three regression models. These are: Linear Regression, Support Vector Regressor \cite{svr}, and XGBoost Regressor \cite{xgboost}. For the baseline prediction, we use the following features: Open, Close, High, Low, and Volume. The target is to predict the next day's Close price using these features. For sentiment analysis applications, we add the sentiment score gathered from the earlier section to the previously mentioned features. We collected 47.3 Million posts/comments from WSB from January 1, 2020, to June 31, 2022, covering both the uncertain times of the global pandemic that had a substantial impact on the stock market, as well as the period surrounding the GME short-squeeze incident that shot WSB to fame. From these 47.3 Million posts/comments, we filtered texts that mention the five stock tickers, which resulted in 2.3 Million posts/comments. We set January 1, 2020, to December 31, 2021, as the training period and January 1, 2022, to June 30, 2022, as the testing period to have an 80:20 split. 


For each of the five stocks, we use texts that mention the respective tickers. Next, we run these relevant texts individually through FinGPT-SA using AIAP and through fine-tuned Bert-Base-Uncased on WSB to gather their sentiment predictions. Finally, we take these predictions and extract each day's sentiment using methods discussed in section \ref{section:sentiment_score_generation}. For the fine-tuned BERT model, we used our proposed qualitative approach (CSBS) detailed in algorithm \ref{algorithm}. For FinGPT-SA, we used the quantitative approach mentioned in equation \ref{eq:baseline_sentiment_score}. The day-to-day stock data for each of the five stocks were collected from Yahoo Finance.

\begin{table*}
\caption{Averaged RMSE and MAE over the three regressors for each asset. Baseline is solely price features, BERT-CSBS represents added sentiment feature from fine-tuned BERT on WSBS using CSBS, and FinGPT-AIAP represents similar feature using AIAP with QuantSS score generation method. BERT and FinGPT represents the default version of the models without CSBS or AIAP integration.}
\centering
\begin{tabular}{ccccccccccccccc}
\toprule
\multirow{2}{*}{Method} & \multicolumn{2}{c}{GME} & \multicolumn{2}{c}{SPY} & \multicolumn{2}{c}{AMC} & \multicolumn{2}{c}{TSLA} & \multicolumn{2}{c}{AAPL} \\ 
\cmidrule(r){2-3} \cmidrule(r){4-5} \cmidrule(r){6-7} \cmidrule(r){8-9} \cmidrule(r){10-11}
 & RMSE & MAE & RMSE & MAE & RMSE & MAE & RMSE & MAE & RMSE & MAE \\ 
\midrule
Baseline & 3.567 & 2.719 & 209.932 & 181.952 & 2.638 & 2.112 & 29.576 & 22.701 & 7.443 & 5.934 \\ 
BERT     & 3.578 & 2.673 & 212.413 & 183.192 & 2.543 & 2.139 & 28.283 & 21.394 & 7.677 & 5.932 \\ 
BERT-CSBS & 3.139 & 2.439 & 212.473 & 184.893 & 2.461 & 1.939 & 27.941 & 21.669 & 7.376 & 5.920 \\ 
FinGPT   & 3.128 & 2.781 & 211.037 & 181.393 & 2.284 & 1.939 & 29.981 & 24.192 & 7.941 & 6.234 \\ 
FinGPT-AIAP & 2.801 & 2.085 & 210.086 & 182.710 & 2.370 & 1.813 & 29.822 & 23.179 & 7.800 & 5.345 \\ 
\bottomrule
\end{tabular}
\label{table:stock_prediction_results}
\end{table*}

\subsection{Stock Prediction Experimental Results}
Table \ref{table:stock_prediction_results} presents the stock prediction results. From the results, it can be inferred that both FinGPT-SA with AIAP and BERT trained on WSBS using our proposed score generation method---CSBS can extract good value from the sentiment analysis of the investor sentiment of WSB in the context of stock price prediction. We compared these models' performances with and without the proposed enhancements and observed better performance overall. The most consistent performance gain can be observed on GME. Right behind GME, we found better RMSE and MAE for AMC and TSLA in that order. On average, we do not see any noteworthy improvement on SPY and AAPL. This discrepancy can be attributed to a couple of things. First, AAPL has a significantly lower number of samples (almost 6.89 times lower than GME) available for sentiment analysis. Second, SPY is a stock index representing a large portion of the stock market. Even though it is established that a myriad of things control the stock prices for each stock other than just the price and sentiment feature from WSB, it is evidently more relevant for a large stock index such as S\&P 500. Despite this, we still observed the best RMSE and MAE scores on SPY from FinGPT and BERT-WSB. 

Aside from this, the improvement in stock prediction with the help of sentimental features from the texts of WSB, especially for GME and AMC, is of little surprise. The reason can be attributed to the GME Short-squeeze event, which opened a perpetual floodgate that flooded academic and financial circles with analysis on WSB, and GME and AMC were the top two most discussed individual stocks in WSB during the timeframe in which we conducted our experiments.

\subsection{Ablation: Importance of WSBS \& CSBS}
\begin{table*}[htbp]
\caption{Stock-wise average RMSE (and respective MAE in trailing parantheses) percentage improvement over all regressors for different score generation methods and sentiment prediction models.}
\centering
\begin{tabular}{cccccccc}
\toprule
\multirow{2}{*}{SA Model}   & \multirow{2}{*}{Method} & \multicolumn{5}{c}{Stock} & \multirow{2}{*}{Average} \\ \cline{3-7} 
                            &            & \rule{0pt}{2.5ex}GME & SPY & AMC & TSLA & AAPL & \\ \midrule
\multirow{1}{*}{FinGPT-AIAP}     & QuantSS   & 13.34 (14.67) & 1.37 (1.16) & 8.29 (11.16) & 0.67 (-0.57) &  -2.18 (-3.59)  & 4.30 (4.57) \\ \midrule
\multirow{2}{*}{BERT-WSBS}   & CSBS       & 9.75 (8.94)   & -0.32 (-0.07)   & 3.79 (4.54)   & 3.78 (2.82) & -0.07 (-0.87)  & 3.39 (3.07) \\
                            & QuantSS   & 0.43 (-0.15)   & -0.70 (-1.33)   & 7.54 (8.43) & 3.15 (2.39) & -3.08 (-5.54) & 1.47 (0.76)   \\ \midrule
\multirow{2}{*}{BERT-FPB}   & CSBS       & -0.43 (-2.21)   & 0.18 (0.44)   & 5.21 (6.88)   & 4.49 (3.12) & 3.29 (2.00)   & 2.55 (2.05)   \\ 
                            & QuantSS   & 12.89 (13.47)   & -0.80 (-1.19)   & 4.81 (6.25)   & -1.93 (-3.54) & -3.00 (-3.60)   & 2.39 (2.28)  \\ \midrule
\multirow{2}{*}{BERT-SST5}  & CSBS       & 0.51 (-1.51)   & 0.45 (0.85)  & 7.94 (11.96)   & -0.65 (-1.27) & 5.28 (4.13)   & 2.71 (2.83)  \\ 
                            & QuantSS   & 0.83 (0.47)   & -2.33 (-3.07)   & 6.66 (10.70)   & -3.44 (-5.05) & -7.70 (-7.66)   & -1.20 (-0.92)  \\\bottomrule

\end{tabular}

\label{table:rmse_stock_prediction_ablation}
\end{table*}

Next, we try to understand if our sentiment scoring method, CSBS, can extract more value out of WSB's sentiment compared to the quantitative scoring method. Additionally, we test the importance of fine-tuning on WSBS, i.e., we take BERT-base-uncased and fine-tune it on two other sentiment prediction datasets and compare the results. We chose a financial-news-based sentiment analysis dataset (Financial Phrase Bank: FPB) and a non-financial/generic sentiment analysis dataset (Stanford Sentiment Treebank 5: SST-5) to fine-tune the classifier. FPB is a three-class (positive/negative/neutral) sentiment classification dataset for finance, and SST-5 is a 5 class generic text dataset \cite{malo2014good,sst}. To adapt to the score generation methods, we convert the SST-5 dataset into SST-3 by categorizing the ``very positive" and ``positive" classes together in one class, and ``very negative" and ``negative" together in one class. These three datasets give us three different versions of BERT (BERT-WSBS, BERT-FPB, BERT-SST5) fine-tuned on WSBS, FPB, and SST-5 datasets, respectively. For each of these three models, we try the two score generation methods and use them as regression features for the same three regressors we used previously. 

In table \ref{table:rmse_stock_prediction_ablation}, we present the average RMSE and MAE improvements for each stock across all the regressors. The improvements represent stock prediction improvement by using additional sentiment features over baseline price features. The results clearly indicate that fine-tuning on WSB had the most advantage over the other datasets. Moreover, our sentiment scoring method can extract more value out of models that were not trained on WSB's texts compared to the quantitative approach.

\section{Relevant Studies}
\label{section: relevant studies}

In this multi-faceted interdisciplinary research centered on financial sentiment analysis, we touch upon several important topics. We categorized these topics into the following groups and discussed relevant works in each topic.

\subsection{Financial Sentiment Analysis}
Financial sentiment analysis efforts can be mainly divided into three groups. First, the pre-machine learning efforts relied on lexical approaches with dictionaries and word counts ~\cite{stone1966general,KEARNEY2014171,henry2006market,henry2008investors,loughran2011liability,i_just_like}. Second, the pre-transformers efforts, where sentiment analysis models were built upon machine-learning \cite{mccallum1996bow,antweiler2004all,li2010information,huang2014evidence} and deep-learning structures such as RNN, LSTM etc \cite{KRAUS201738,8334488,lutz2018sentencelevel,severyn2015twitter,sohangir2018big}. The third group of efforts can be attributed to Transformer-based \cite{vaswani2017attention} models, which took over traditional deep-learning models for financial textual analysis for their ability to overcome sequential token processing employing better contextual understanding capacity.

Transformer-based models for financial tasks have been widely used since fine-tuning the Pre-trained language models became popular for domain-oriented tasks \cite{devlin2018bert,liu2020k}. One of the most noteworthy efforts in this space has been by Araci with FinBERT \shortcite{araci2019finbert}. They took a fine-tuning approach to producing a financial sentiment analysis model by capitalizing on BERT's pre-trained architecture. Two other works with the same name (FinBERT) have also been conducted to accomplish financial text processing \cite{liu2021finbert,yang2020finbert}. 
\cite{liu2021finbert} adopted a rather drastic approach by training the base architecture from the ground up using six tasks instead of the original two tasks upon which BERT was trained. Araci's FinBERT has also been extended and utilized in more specialized financial sentiment analysis tasks in \cite{gossi2023finbert}, where they simplify the task by eliminating sentence clauses appearing before select conjunctions.

\subsection{Financial Large Language Models}
With the popularity of generative Large Language Models (LLMs), which are exponentially larger transformer models compared to BERT and other pre-trained models with under a billion parameters, came specialized LLMs in finance, also referred to as FinLLMs \cite{li2023large}. FinLLMs have been grouped into two kinds by Li et al.: Fine-tuned LLMs (FinMA, FinGPT, FinLLaMa etc.) and pre-trained from scratch LLMs (BloombergGPT, Fin-T5 etc.) \cite{li2023large,xie2023pixiu,yang2023fingpt,wu2023bloomberggpt,Fin-LLAMA,lu2023bbt}. The distinction here is almost similar to the distinction between fine-tuned FinBERT and pre-trained FinBERT from scratch; the first builds on generic LLM using financial text, and the other is pre-trained from scratch using financial/mixed corpora. 

However, the primary issue remained with many of the FinLLMs in performing tasks like financial sentiment analysis without further intervention due to their core ``casual language modeling" objective of simply predicting the next token \cite{zhang2023enhancing}, which produced Instruction-tuned FinLLMs that can understand human instructions. Instruct-FinGPT is one such model that takes a generic LLM and is instruction-tuned on samples that consist of instructions and answers. With instruction-tuned LLMs, the next challenge came in understanding if the instructions we provide to the LLMs are understood properly by these LLMs. From few-shot learning introduced in \cite{brown2020language} to chain-of-thought prompting in \cite{wei2022chain}, it has been observed that LLMs retain the capacity to learn during inference from their instructions, but have not been fully explored in FinLLMs. This study, however, focuses on enhancing financial sentiment analysis by exploring effective prompting methods in LLMs.



\section{Conclusion}
\label{section: conclusion}
The key to success behind prompt engineering relied upon the alignment of what humans know and what we can make them understand with these instructions. This study extends this idea by observing the financial sentiment analysis task from a fresh perspective. We guide a team of annotators with a well-constructed instruction towards a financial sentiment analysis task on texts from WallStreetBets and try to see if LLMs, capable of understanding instructions, can align its reasoning with the annotators using the same instruction. Annotators' instruction assisted prompting (AIAP) dramatically improves LLMs' sentiment analysis capabilities, indicating that the current benchmarking efforts involve highly subjective labeling, which result in improper evaluation. Furthermore, we explore the application of this new-found sentiment prediction prowess in stock price prediction for a well-rounded analysis. We also look at small-scaled encoder-based models for the sentiment analysis application study using a novel sentiment scoring technique and our annotated dataset as a fine-tuning source. We observed the importance of each of these components in predicting stock prices.

\bibliographystyle{ACM-Reference-Format}
\bibliography{main}
\end{document}